\documentclass[
]{ceurart}

\usepackage{listings}
\usepackage{tabularx}
\usepackage{caption}
\usepackage{float}
\usepackage{graphicx}
\begin{document}

\copyrightyear{2024}
\copyrightclause{Copyright for this paper by its authors.
  Use permitted under Creative Commons License Attribution 4.0
  International (CC BY 4.0).}

\conference{Woodstock'22: Symposium on the irreproducible science,
  June 07--11, 2022, Woodstock, NY}

\title{Breaking Down Financial News Impact: A Novel AI Approach with Geometric Hypergraphs}





\author[1,2]{Anoushka Harit}[%
orcid=0000-0002-8185-4790,
email=anoushkaharit3@gmail.com,
]
\cormark[1]
\fnmark[1]

\address[1]{Department of Computer Science, Durham University, Stockton Rd, Durham, DH1 3LE}

\address[2]{Cancer Research UK Cambridge Institute, University of Cambridge Li Ka Shing Centre, Robinson Way, Cambridge, CB2 0RE}

\author[3]{Zhongtian Sun}[%
email=zs440@cam.ac.uk,
]
\fnmark[1]

\address[3]{Department of Department of Applied Mathematics and Theoretical Physics, University of Cambridge, Wilberforce Rd, Cambridge, CB3 0WA}

\author[3]{Jongmin Yu}[%
email=jy522@cam.ac.uk,
]

\author[1]{Noura Al Moubayed}[%
email=noura.al-moubayed@durham.ac.uk,
]
\cortext[1]{Corresponding author.}
\fntext[1]{These authors contributed equally.}

\begin{abstract}
 In the fast-paced and volatile financial markets, accurately predicting stock movements based on financial news is critical for investors and analysts. Traditional models often struggle to capture the intricate and dynamic relationships between news events and market reactions, limiting their ability to provide actionable insights. This paper introduces a novel approach leveraging Explainable Artificial Intelligence (XAI) through the development of a Geometric Hypergraph Attention Network (GHAN) to analyze the impact of financial news on market behaviours. Geometric hypergraphs extend traditional graph structures by allowing edges to connect multiple nodes, effectively modelling high-order relationships and interactions among financial entities and news events. This unique capability enables the capture of complex dependencies, such as the simultaneous impact of a single news event on multiple stocks or sectors, which traditional models frequently overlook.

By incorporating attention mechanisms within hypergraphs, GHAN enhances the model's ability to focus on the most relevant information, ensuring more accurate predictions and better interpretability. Additionally, we employ BERT-based embeddings to capture the semantic richness of financial news texts, providing a nuanced understanding of the content. Using a comprehensive financial news dataset, our GHAN model addresses key challenges in financial news impact analysis, including the complexity of high-order interactions, the necessity for model interpretability, and the dynamic nature of financial markets. Integrating attention mechanisms and SHAP values within GHAN ensures transparency, highlighting the most influential factors driving market predictions.
Empirical validation demonstrates the superior effectiveness of our approach over traditional sentiment analysis and time-series models. Our framework not only improves prediction accuracy but also provides detailed insights into how financial news impacts different market sectors. This comprehensive analysis empowers investors and analysts with deeper, actionable insights, setting the stage for future research in applying advanced AI methodologies to financial analysis and ultimately advancing the field.
\end{abstract}


\maketitle

\section{Introduction}
In recent years, financial technology (FinTech) has revolutionized the stock market, making it more accessible yet increasingly complex \cite{baek2018modaugnet}. While conventional stock analysis primarily focuses on predicting stock prices, there is a growing demand for sophisticated methods to recommend the most profitable stocks \cite{sun2023money}. This shift reflects investors' primary interest: identifying stocks that can bring higher returns in the future \cite{cao2003support}.
Traditional approaches to stock prediction, such as Auto Regression-based methods \cite{chen2015lstm} and Support Vector Regression \cite{chen2018incorporating}, often fall short of capturing the dynamic, non-linear nature of financial markets. Deep learning methods, including Recurrent Neural Networks \cite{cho2014learning, sun2022unimodal, sun2024robustness} and attention-based models \cite{sun2021generative, sun2022contrastive, sun2023adaptive, sun2023rewiring}, have shown promise in learning sequential patterns from longitudinal data. However, these models typically overlook the complex, multi-dimensional relationships between stocks, sectors, and crucial external factors such as financial news.
Recent advancements in graph neural networks have opened new avenues for modelling stock market dynamics \cite{ding2019modeling, sun2023money}. These approaches can capture relationships between stocks. However, they often rely on pre-defined connections and struggle to incorporate real-time news information, limiting their ability to adapt to the rapidly changing financial landscape \cite{feng2019enhancing}. Moreover, existing models fail to fully leverage the rich, textual data from financial news sources, which can provide valuable insights into market trends and company performance \cite{feng2019temporal}.
To address these challenges, we propose the Geometric Hypergraph Attention Network (GHAN), a novel deep learning model designed to capture the complex, high-order relationships in financial markets while integrating real-time news data. Our work makes several key contributions:

\begin{enumerate}
    \item Hypergraph Modeling with News Integration: GHAN leverages hypergraph structures to model multi-entity interactions, including stocks, sectors, and news events [14]. This allows us to capture complex interactions between multiple stocks and news items simultaneously, a significant advancement over traditional graph-based models.
    \item Real-time Financial News Processing: We incorporate data from major financial news sources like Twitter Finance news\footnote[2]{https://huggingface.co/datasets/zeroshot/twitter-financial-news-sentiment}, enabling GHAN to capture the immediate impact of news on stock performance. This real-time integration of textual data sets our model apart from existing approaches that rely solely on historical price data.
    \item Geometric Information Integration: We incorporate geometric information through positional encodings, enabling GHAN to capture implicit spatial and temporal relationships within the financial ecosystem and news landscape.
    \item Latent Relationship Learning: GHAN learns latent interactions between stocks, sectors, and news events without relying on pre-defined relationships. This approach enables the model to adapt to changing market dynamics and uncover hidden influences that may not be apparent through traditional analysis.
    \item Explainability: GHAN incorporates attention mechanisms that allow for the interpretability of the model's decisions, addressing the crucial need for transparency in financial prediction models, especially when incorporating new data.
\end{enumerate}
In this paper, we demonstrate how GHAN effectively recommends the most profitable stocks by modelling these hierarchical correlations within the market and leveraging real-time financial news. Our experiments on real-world financial datasets and news feeds show that GHAN significantly outperforms state-of-the-art methods, offering investors a powerful tool for making informed decisions in the complex and fast-paced world of stock trading.By addressing the limitations of existing approaches and introducing novel techniques for integrating financial news into market modelling, GHAN represents a significant step forward in the field of stock recommendation and analysis.

\section{Related Work}
The analysis of financial markets and the prediction of stock movements based on news events is a complex, multifaceted problem that intersects several research domains. Our work draws inspiration from and builds upon advancements in financial market prediction, graph-based deep learning, hypergraph neural networks, explainable AI, and geometric deep learning. This section provides an overview of the key developments in these areas that form the foundation of our research. We begin by discussing traditional and modern approaches to financial market prediction, followed by the application of graph neural networks in finance. We then explore the emerging field of hypergraph neural networks and their potential for modelling complex financial ecosystems. The growing importance of model interpretability in finance leads us to review recent work in explainable AI. Finally, we examine the field of geometric deep learning and its applications in financial modelling. Our research builds upon and extends several key areas in the fields of financial technology, graph neural networks, and explainable AI.

\begin{enumerate}
    \item \textbf{Stock Market Prediction with News Data}:
    The integration of financial news data for stock market prediction has been an active area of research. Hu et al. [1] proposed a deep learning framework that listens to chaotic stock market news to predict stock trends. Their model captures both short-term and long-term dependencies in financial news. Similarly, Vargas et al. [2] developed a deep learning model that combines both textual and technical analysis for stock market prediction.
    Xie et al. [3] introduced a method using semantic frames to predict stock price movement, demonstrating the importance of capturing semantic information from financial news. These works highlight the potential of incorporating news data into stock prediction models but cannot often capture complex, multi-entity relationships.

    \item\textbf{Graph-Based Approaches in Finance}:
     Graph-based models have shown promising results in capturing relationships between financial entities. \cite{chen2018incorporating} incorporated corporation relationships via graph convolutional neural networks for stock price prediction. \cite{sawhney2021stock} proposed a temporal attention-augmented graph convolutional network for stock recommendation. \cite{sun2023money}  demonstrate the effectiveness of integrating auxiliary information via GNNs before using RNNs for temporal studies. However, these approaches are limited by their use of simple graph structures, which cannot fully capture the complex, multi-entity interactions present in financial markets. 

     \item\textbf{Hypergraph Neural Network}:
     Hypergraph neural networks allow for modelling higher-order relationships beyond pairwise interactions. \cite{feng2019hypergraph} introduced hypergraph attention networks for social recommendation systems. In the financial domain, \cite{xin2024multiple} applied hypergraph convolution to model multi-faceted relationships between stocks, though their approach did not incorporate geometric information or explainable AI techniques.

     \item \textbf{Explainable AI Finance}:
     As AI models become more complex, the need for explainability has grown, especially in high-stakes domains like finance. \cite{lundberg2017unified} introduced SHAP (SHapley Additive exPlanations) values as a unified measure of feature importance. In the context of financial forecasting, \cite{nagy2024interpretable} used SHAP values to explain the predictions of a deep learning model for stock price movement. Our work bridges these areas by introducing a Geometric Hypergraph Attention Network that incorporates positional encodings to capture spatial and temporal relationships in financial data and news. We also integrate SHAP values for model explainability, addressing the critical need for interpretable AI in financial decision-making.
     
\end{enumerate}

\section{Problem Statement}
Investors often have sufficient capital but lack the insights needed to make informed decisions about which stocks are most promising for investment. To address this challenge, we aim to analyze the impact of financial news on stock market behaviours using a Geometric Hypergraph Attention Network (GHAN). Our goal is to capture the complex, multi-dimensional relationships between stocks and news events and predict their influence on stock movements.Let $S=\left\{s_1, s_2, \ldots, s_n\right\}$ be the set of $n$ stocks, and $N=\left\{n_1, n_2, \ldots, n_m\right\}$ be the set of $m$ news events. We represent this financial ecosystem as a hypergraph $\mathcal{H}=(V, E)$, where $V=S \cup N$ is the set of nodes (stocks and news events), and $E$ is the set of hyperedges connecting multiple stocks and news events.\\

For each stock $s_i \in S$, we define its feature vector $\mathbf{x}_i \in \mathbb{R}^d$, which includes relevant financial indicators such as price-based, momentum, and sentiment indicators.

For each news event $n_j \in N$, we define:

\begin{enumerate}
  \item Its feature vector $\mathbf{y}_j \in \mathbb{R}^u$, derived from natural language processing (NLP) techniques applied to the news content, including sentiment analysis, topic modelling, and named entity recognition (NER).
  \item A positional encoding $\mathbf{p}_j \in \mathbb{R}^p$, which captures the temporal and contextual information of the news event, such as publication time, source credibility, and topic relevance.
\end{enumerate}
The positional encoding $\mathbf{p}_j$ is crucial as it provides the geometric information in our GHAN model, allowing us to capture the 'spatial' and temporal relationships within the news landscape and their impact on stocks. Our GHAN model computes attention coefficients $\alpha_{i j}$ for the importance of node $i$ (stock or news) to hyperedge $j$ :
$$
\alpha_{i j}=\frac{\exp \left(\operatorname{LeakyReLU}\left(\mathbf{a}^T\left[\mathbf{W}_x \mathbf{x}_i\left\|\mathbf{W}_y \mathbf{y}_j\right\| \mathbf{W}_p \mathbf{p}_j\right]\right)\right)}{\sum_{k \in e_j} \exp \left(\operatorname{LeakyReLU}\left(\mathbf{a}^T\left[\mathbf{W}_x \mathbf{x}_k\left\|\mathbf{W}_y \mathbf{y}_j\right\| \mathbf{W}_p \mathbf{p}_j\right]\right)\right)}
$$
where $\mathbf{W}_x, \mathbf{W}_y$, and $\mathbf{W}_p$ are learnable weight matrices, $\mathbf{a}$ is an attention vector, and $\|$ denotes concatenation. Note that the positional encoding $\mathbf{p}_j$ is only applied to news events.The model updates stock representations as:
$$
\mathbf{x}_i^{\prime}=\sigma\left(\sum_{j \in \mathcal{N}(i)} \alpha_{i j} \mathbf{W}\left[\mathbf{y}_j \| \mathbf{p}_j\right]\right)
$$
where $\mathcal{N}(i)$ is the set of news events connected to stock $i$, and $\sigma$ is an activation function.\\

Given historical data $D=\left\{\left(\mathcal{H}_t, \mathbf{R}_t\right) \mid t=1, \ldots, T\right\}$, where $\mathcal{H}_t$ is the hypergraph state at time $t$ and $\mathbf{R}_t$ is the corresponding stock return vector, our objective is to learn a function $f$ that predicts future stock returns:
$$
\hat{\mathbf{R}}_{t+1}=f\left(\mathcal{H}_t\right)
$$

We aim to minimize the prediction error while maintaining interpretability through the attention mechanisms and the news-based geometric embeddings. This approach captures the intricate effects of news events on multiple stocks, modelling the temporal and contextual aspects of financial news to understand their market impact. By leveraging positional encodings from news content, we effectively represent the 'spatial' relationships within the news landscape. This methodology provides deep insights into the complex interactions between news and stock movements, significantly enhancing stock return forecasts' predictive accuracy and interpretability.

\section{Methodology}
\subsection{Overview}
we integrate S\&P 500 stock data with Twitter Financial News Sentiment data to predict stock movements using a Geometric Hypergraph Attention Network (GHAN). Initially, the S\&P 500 stock data is preprocessed to compute moving averages, while financial news tweets undergo preprocessing and are embedded using FinBERT to capture their contextual information. We then construct a hypergraph where nodes represent stocks and tweets, and hyperedges connect tweets to the stocks mentioned within them, effectively modelling high-order interactions between the data points. The basic framework is as follows:
\begin{center}
    \includegraphics[scale=1]{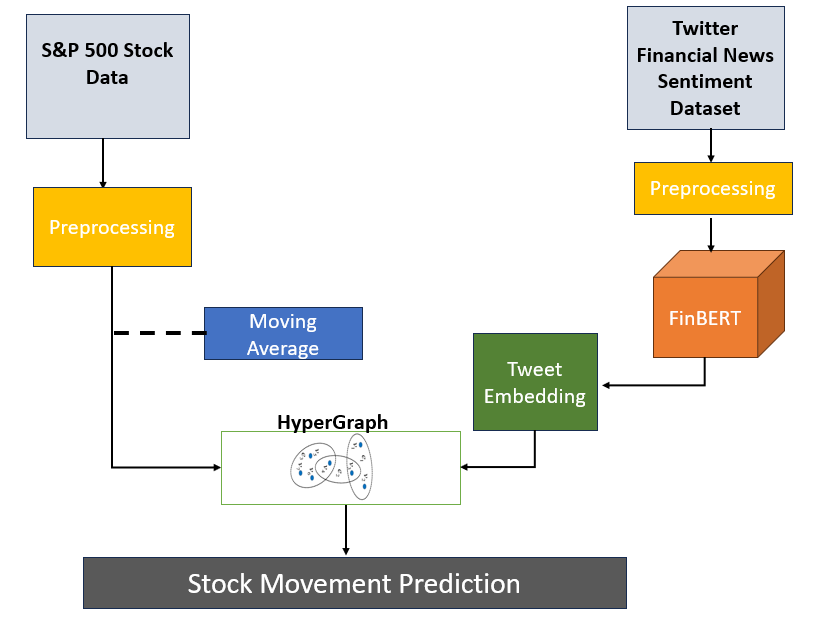}
    
    Fig 1: GHAN Basic Workflow
\end{center}

The implementation of the workflow will be further detailed in the following subsections, showcasing how our GHAN model effectively integrates geometric embeddings, FinBERT tweet embeddings, and hypergraph structures to enhance the accuracy and precision of financial sentiment analysis and stock movement prediction. Additionally, we will demonstrate the use of SHAP values to provide explainable insights into the model's predictions.

\subsection{Hypergraph-Level Modelling}
The hypergraph-level modelling is the core component of our GHAN approach, designed to capture the intricate and high-order relationships between stocks and financial news events, unlike traditional graph structures that only model pairwise relationships, hypergraphs allow us to represent complex, multi-entity interactions that are prevalent in financial markets.
\subsubsection{Hypergraph Structure and Feature Extraction}

We represent the financial ecosystem as a hypergraph $\mathrm{H}=(\mathrm{V}, \mathrm{E})$, where V is the set of nodes and $E$ is the set of hyperedges. This structure is crucial for our analysis because:

\begin{enumerate}
    \item  Nodes (V) represent stocks and news events, allowing for a unified representation of different entity types.
    \item Hyperedges (E) can connect multiple nodes, enabling us to model complex relationships such as the simultaneous impact of a single news event on multiple stocks or sectors.
\end{enumerate}

The incidence matrix $\mathrm{H} \in \mathrm{R}^{\wedge}(|\mathrm{V}| \times|\mathrm{E}|)$ provides a mathematical representation of these connections, where $\mathrm{H}\left(\mathrm{v}_{\mathrm{i}}, e_{\mathrm{j}}\right)=1$ if node $v_i$ is part of hyperedge $e_j$, and 0 otherwise. This matrix is fundamental to our subsequent attention mechanisms and information propagation processes.

For feature extraction, we carefully select a set of relevant attributes for both stocks and news events. Stock features include standard financial indicators, while news features are derived from advanced NLP techniques. This diverse feature set ensures that our model comprehensively views quantitative market data and qualitative news information.

\subsubsection{Geometric Embedding}
The incorporation of geometric embeddings is a key innovation in our GHAN model. By adding positional encodings $P_V$ and $P_E$ to node and hyperedge features respectively, we imbue our model with a sense of 'spatial' relationships within the financial landscape. This is particularly important because:

\begin{enumerate}
    \item It allows the model to capture implicit relationships between entities based on their 'position' in the financial ecosystem.
    \item It provides a mechanism for the model to learn and utilize structural information that may not be explicitly present in the feature vectors.
\end{enumerate}
 The combination of original features with these geometric embeddings as $\mathbf{X} \leftarrow \mathbf{X}+\mathbf{P}_V$ and $\mathbf{E} \leftarrow \mathbf{E}+\mathbf{P}_E$
 creates rich, context-aware representations that serve as the foundation for our attention mechanisms.
\subsubsection{BERT Embedding}
 We use the FinBERT model \cite{araci2019finbert}, which is specifically fine-tuned for financial text. This model leverages the capabilities of BERT while being adapted to understand the nuances of financial language. Each tweet is tokenized using the BERT tokenizer, which converts the text into a format that the FinBERT model can process. This involves breaking down the text into word pieces and adding special tokens like [CLS]. We pass the tokenized tweets through the FinBERT model to obtain embeddings. Specifically, we extract the embeddings from the last hidden layer corresponding to the [CLS] token, which represents the aggregated information of the entire tweet.
 Let $t$ represent a tweet. The embedding $\operatorname{FinBERT}(t)$ is obtained as follows:
$$
\operatorname{FinBERT}(t)=\operatorname{FinBERT} \text { model }(\operatorname{Tokenize}(t))[\operatorname{CLS}]
$$
representation of the tweet $t$.We combine the FinBERT-generated tweet embeddings with geometric and positional embeddings as follows:
\begin{equation}
e_{\text {final }}=g_e+p_e+\operatorname{FinBERT}\left(t_e\right)
\end{equation}
where $\operatorname{FinBERT}\left(t_e\right)$ is the embedding of tweet $t_e$ obtained from the FinBERT model.

\subsubsection{Hypergraph Attention Network}
Our multi-level attention mechanism is designed to model the complex, hierarchical nature of interactions in financial markets. It operates on two levels: Node-Level Attention and Hyperedge-Level Attention. 

\begin{enumerate}
    \item Node Level Attention: This mechanism allows the model to weigh the importance of different nodes within each hyperedge. Mathematically, for a node $i$ and a hyperedge j, we compute the attention coefficients $\alpha_{i j}$ as follows \\
    \begin{center}
     $\alpha_{i j}=\frac{\exp \left(\operatorname{LeakyReLU}\left(\mathbf{a}^2\left[\mathbf{W}\left(\mathbf{x}_i+\mathbf{p}_i\right) \| \mathbf{W}\left(\mathbf{e}_j+\mathbf{p}_j\right)\right]\right)\right)}{\sum_{k \in e_j} \exp \left(\operatorname{LeakyReLU}\left(\mathbf{a}^T\left[\mathbf{W}\left(\mathbf{x}_k+\mathbf{p}_k\right) \| \mathbf{W}\left(\mathbf{e}_j+\mathbf{p}_j\right)\right]\right)\right)}$
     \end{center}

     where

     \begin{enumerate}
         \item  $\mathbf{W}$ is a weight matrix.
         \item $\mathbf{x}_i$ is the feature vector of node $i$.
         \item $\mathbf{e}_j$ is the feature vector of hyperedge $j$.
         \item $\mathbf{a}$ is the attention mechanism weight vector.
         \item $\|$ denotes concatenation.
   
     \end{enumerate}
By employing this node-level attention mechanism, our model captures the varying importance of different stocks and news events within each market sector or news category.\\

\item Hyper-Edge Level Attention: This mechanism enables the model to assess the relative importance of different hyperedges for each node. The attention coefficient $\beta_{j i}$ is computed as follows

\begin{center}
    $\beta_{j i}=\frac{\exp \left(\operatorname{LeakyReLU}\left(\mathbf{b}^T\left[\mathbf{W}\left(\mathbf{e}_j+\mathbf{p}_j\right) \| \mathbf{W}\left(\mathbf{x}_i+\mathbf{p}_i\right)\right]\right)\right)}{\sum_{l \in \mathcal{N}(i)} \exp \left(\operatorname{LeakyReLU}\left(\mathbf{b}^T\left[\mathbf{W}\left(\mathbf{e}_l+\mathbf{p} l\right) \| \mathbf{W}\left(\mathbf{x}_i+\mathbf{p}_i\right)\right]\right)\right)}$
\end{center}
where 
\begin{enumerate}
    \item $\mathbf{W}$ is a weight matrix.
    \item $\mathbf{x}_i$ is the feature vector of node $i$.
    \item $\mathbf{e}_j$ is the feature vector of hyperedge $j$.
    \item $\mathbf{b}$ is another attention mechanism weight vector.
    \item $\mathcal{N}(i)$ is the set of hyperedges connected to node $i$.
\end{enumerate}

\end{enumerate}
The attention coefficients ($\alpha_{i j}$ and $\beta_{j i}$) are computed using learnable parameters, allowing the model to adaptively focus on the most relevant connections as it processes the data. This formulation allows the model to capture complex interactions by considering both the importance of nodes within hyperedges and the importance of hyperedges for each node, providing a rich, hierarchical representation of the financial market structure.

\subsection{Information Aggregation}
The final step in our hypergraph-level modelling is the aggregation of information. This process updates both node and hyperedge representations based on the learned attention weights which is mathematically formulated as

\begin{enumerate}
    \item Node Update :
     The node update ($\mathbf{x}_i^{\prime}$) incorporates information from connected hyperedges
     \begin{center}
       \begin{equation}
x_i^{\prime}=\sigma\left(\sum_{j \in N(i)} \alpha_{i j} W\left(e_j+p_j+\operatorname{FinBERT}\left(t_j\right)\right)\right)
\end{equation}
     \end{center}
     \item Hyperedge Update: the hyperedge update ($\mathbf{e}_j^{\prime}$) aggregates information from its constituent nodes.
     \begin{center}
     \begin{equation}
e_j^{\prime}=\sigma\left(\sum_{i \in e_j} \beta_{j i} W\left(x_i+p_i+\operatorname{FinBERT}\left(t_i\right)\right)\right)
\end{equation}
     \end{center}
\end{enumerate}
By leveraging the hypergraph structure and incorporating BERT embeddings, our GHAN model can effectively capture and utilize the complex, multi-dimensional relationships present in financial markets. This sophisticated approach to modelling market dynamics provides a strong foundation for subsequent tasks such as predicting the impact of news on stock movements and offering explainable insights into market behaviours. The integration of geometric embeddings, positional encodings, and contextual information from FinBERT enables our model to accurately interpret financial news and its influence on stock prices, thus enhancing prediction accuracy and interoperability.

\subsection{SHAP for Model Explainability}
To interpret the predictions made by the GHAN model, we utilize \cite{lundberg2017unified} SHAP (SHapley Additive exPlanations) method, which provides a unified measure of feature importance. For a given prediction, the SHAP value $\phi_i$ for feature $\imath$ is computed as:
$$
\phi_i=\sum_{S \subseteq N \backslash\{i\}} \frac{|S|!(|N|-|S|-1)!}{|N|!}[f(S \cup\{i\})-f(S)]
$$
where:
$N$ is the set of all features,
$S$ is a subset of features,
and$f(S)$ is the model prediction for the subset $S$.

\section{Experimental Setup}
In this section, we present a comprehensive evaluation of our proposed Geometric Hypergraph Attention Network (GHAN) for analyzing the impact of financial news on stock prices. Our experiments aim to demonstrate the effectiveness and limitations of GHAN compared to several state-of-the-art financial prediction methods.
\subsection{Data Collection and Processing}
Our research focuses on the developed stock market of the United States, specifically analyzing stocks of enterprises included in the S\&P 500 index. We obtain historical price data for stocks in the S\&P 500 Composite index from the Yahoo Finance website \footnote{https://finance.yahoo.com/} . Our dataset includes 450 stocks from the S\&P 500 index after excluding those with insufficient data or trading history. 

$$
\text { Table 1. Statistics of historical price data. }
$$
\begin{tabular}{ccccc}
\hline Index & Stocks & Training Days & Validation Days & Testing Days \\
\hline \multirow{2}{*}{$S\&$ P 500 } & \multirow{2}{*}{450} & $08 / 02 / 2015-23 / 05 /2020$& 24/05/2017-27/03/2019 & 27/03/2019-29/08/2020 \\
& & 1931 days & 672 days & 521days \\
\hline
\end{tabular}\\

The historical price dataset used in this paper is described in detail in Table 1. Instead of using raw price data, we calculate the daily price change rate as input for our model. The rate of change in the price of a stock at time $t$ is calculated by:
$$
R_i^t=\frac{P_i^t-P_i^{(t-1)}}{P_i^{(t-1)}}
$$
where $P_i^t$ and $P_i^{(t-1)}$ are the closing prices of stock i at time t and $\mathrm{t}-1$, respectively. For financial news data, we utilize the Twitter Financial News Sentiment dataset available on Hugging Face \footnote{https://huggingface.co/datasets/zeroshot/twitter-financial-news-sentiment}. This dataset consists of 500,000 financial news tweets with sentiment labels (positive, negative, neutral), specifically designed for sentiment analysis tasks in the financial domain. Our final text dataset comprises 50,000 tweets related to the S\&P 450 stocks.
     
\subsection{Hypergraph Constrcution}
We define nodes as stock and news events and aim to capture the intricate relationships between stocks and news events, facilitating high-order interactions essential for financial analysis.
\begin{enumerate}
    \item \textbf{Node Definition} :
     Let $V=S \cup N$ be the set of nodes, where
     
     $S=\left\{s_1, s_2, \ldots, s_n\right\}$ is the set of $n$ stocks.\\
     $N=\left\{n_1, n_2, \ldots, n_m\right\}$ is the set of $m$ news events.
     
     \item \textbf{Hyperedge Creation}: 
     Let $E=\left\{e_1, e_2, \ldots, e_k\right\}$ be the set of hyperedges, where each hyperedge $e_i \subseteq$ $V$.\\
     
     For each news event $n_j \in N$ , we create a hyperedge $e_i=\left\{n_j\right\} \cup\left\{s_k \mid\right.$ stock $s_k$ is mentioned in news event $\left.n_j\right\}$.This process results in $k$ hyperedges, where $k \leq m$ since some news events may not mention any stocks.

     \item \textbf{Incidence Matrix Construction}:
      Construct the incidence matrix $\mathcal{H} \in \mathbb{R}^{|V| \times|E|}$ where:
\begin{equation}
       \mathcal{H}(v, e)= \begin{cases}1, & \text { if } v \in e \\ 0, & \text { otherwise }\end{cases}
\end{equation}
for $v \in V$ and $e \in E$.The incidence matrix effectively encodes the hypergraph structure, enabling efficient computation of relationships between nodes and hyperedges during model training and inference.
\item \textbf{Final Embedding}:

For each entity $e \in V$ :
Compute geometric embedding $g_e \in \mathbb{R}^{100}$.
Compute positional encoding $p_e \in \mathbb{R}^{100}$.
 Compute and combine the FinBERT-generated tweet to get the final embedding as 
 \begin{equation}
e_{\text {final }}=g_e+p_e+\operatorname{FinBERT}\left(t_e\right)
\end{equation}

\end{enumerate}

\subsection{Mathematical Formulation for Data Integration}
To integrate the Twitter financial news sentiment data with the $S \& P 500$ stock price data, we perform several key steps. First, we define our data sets: let $T$ be the set of tweets, where each tweet $t \in T$ is associated with a sentiment score $s_t$ and a timestamp $\tau_t$ let $S$ be the set of stocks, where each stock $s \in S$ has a corresponding closing price $P_{\mathrm{s}}^t$ at time $t$.Next, we align tweets with stock data. For each stock $s \in S$, we aggregate the tweets mentioning $s$ regularly. Let $T_s^t$ denote the set of tweets related to stock $s$ on day $t$. We then compute the aggregated sentiment score $\bar{s}_s^t$ for stock $s$.

\begin{equation}
\bar{s}_s^t=\frac{1}{\left|T_s^t\right|} \sum_{t \in T_s^t} s_t
\end{equation}
where $\left|T_s^t\right|$ is the number of tweets mentioning stock $s$ on day $t$. Following sentiment aggregation, we construct a feature vector $z_s^t$ for each stock $s$ on day $t$, which includes the daily return $R_s^t$ :
$$
R_s^t=\frac{P_s^t-P_s^{t-1}}{P_s^{t-1}}
$$
the aggregated sentiment score $\bar{s}_{s^{\prime}}^t$ trading volume, and specific technical indicators such as the 20-day moving average (MA20), the 50-day moving average (MA50), the Relative Strength Index (RSI), and the Moving Average Convergence Divergence (MACD).

For hypergraph construction, we define the set of nodes $V=S \cup T$, where $S$ is the set of stock nodes and $T$ is the set of tweet nodes. We then define the set of hyperedges $E$, where each hyperedge $e \in E$ connects a tweet node $t$ to the corresponding stock nodes $\{s \in S \mid t$ mentions $s\}$. Finally, we construct the incidence matrix $\mathcal{H} \in \mathbb{R}^{|V| \times|E|}$, where:
$$
\mathcal{H}(v, e)= \begin{cases}1, & \text { if } v \in e \\ 0, & \text { otherwise }\end{cases}
$$

This matrix encodes the hypergraph structure, indicating the connections between nodes and hyperedqes. The integrated dataset $\left\{z_s^t \mid s \in S, t \in T\right\}$ combines quantitative stock price data and qualitative sentiment data from tweets. This dataset serves as the input for the Geometric Hypergraph Attention Network (GHAN), enabling the model to leverage both market sentiment and historical stock performance for accurate financial analysis and prediction.

\subsection{Model Training}
Our Geometric Hypergraph Attention Network (GHAN) is implemented using the PyTorch framework, leveraging its robust support for dynamic computation graphs and efficient tensor operations. The model is optimized with the Adam optimizer, featuring a learning rate of $5 \times 10^{-4}$ and a weight decay of $5 \times 10^{-5}$ to prevent overfitting. Training is conducted with a batch size of 32 over 100 epochs, incorporating a dropout rate of 0.5 at the end of each layer to mitigate overfitting.

We use the LeakyReLU activation function in the hidden layers to introduce nonlinearity and address the vanishing gradient problem. Our attention mechanism includes node-level attention coefficients computed as:
$$
\alpha_{i j}=\operatorname{softmax}_j\left(e_{i j}\right)
$$
and hyperedge-level attention coefficients as:
$$
\beta_{j i}=\operatorname{softmax}_i\left(f_{j i}\right)
$$

The node update is performed with:
$$
x_i^{\prime}=\sigma\left(\sum_{j \in N(i)} \alpha_{i j} W\left(e_j+p_j\right)\right)
$$
and the hyperedge update follows:
$$
e_j^{\prime}=\sigma\left(\sum_{i \in e_j} \beta_{j i} W\left(x_i+p_i\right)\right)
$$
To enhance the interpretability of our model, we integrate SHapley Additive exPlanations (SHAP). SHAP values help to understand the contribution of each feature to the model's predictions. SHAP values are derived from game theory and provide a unified measure of feature importance. The SHAP value $\phi_i$ for a feature $i$ is given by:

$$
\phi_i=\sum_{S \subseteq F \backslash\{i\}} \frac{|S|!(|F|-|S|-1)!}{|F|!}(v(S \cup\{i\})-v(S))
$$

where $F$ is the set of all features, $S$ is a subset of features, and $v(S)$ is the model prediction given the features in subset $S$. This equation effectively distributes the total prediction among the features, attributing an importance value to each.

\subsection{Basline}
\begin{enumerate}
    \item \textbf{LSTM with Attention}:
    This model uses a Long Short-Term Memory (LSTM) network combined with an attention mechanism to capture temporal dependencies in the financial news data. The attention mechanism helps the model focus on the most relevant parts of the input sequence when making predictions.

    \item \textbf{Graph Convolutional Network (GCN)}:
    GCNs are designed to work directly with graph-structured data. In this baseline, we apply GCN to the financial data graph, where nodes represent stocks and edges represent relationships based on co-occurrence in news events. The GCN aggregates information from neighbouring nodes to make predictions.

    \item\textbf{BERT-based Text Classification}:
    We use a BERT-based model fine-tuned for text classification tasks. This model processes financial news text to predict stock movements, leveraging the powerful contextual embeddings generated by BERT to understand the sentiment and implications of the news content.

    \item \textbf{ML-GAT (Multi-level Graph Attention Network)}:
    ML-GAT applies attention mechanisms at multiple levels within a graph structure, allowing the model to weigh the importance of different nodes and edges differently. This baseline is particularly useful for capturing complex interactions in the financial news and stock data.

    \item \textbf{Fine-BERT}:
    A variant of BERT, Fine-BERT is fine-tuned specifically on financial news data. This model aims to capture the nuanced language and context-specific to financial markets, enhancing the prediction accuracy for stock movements.
\end{enumerate}

\subsection{Evaluation and Analysis}
To compare the performance of the proposed Geometric Hypergraph Attention Network (GHAN) with benchmark models, we use common metrics for evaluating classification and profitability. Stock trend prediction is a typical classification task, so we selected two widely used evaluation indicators: accuracy and F1-score. The calculation formulas are as follows:

\begin{equation}
\begin{aligned}
& \text { Accuracy }=\frac{T P+T N}{T P+T N+F P+F N} \\
& \text { F1-score }=2 \times \frac{\text { Recall } \times \text { Precision }}{\text { Recall }+ \text { Precision }}
\end{aligned}
\end{equation}

where $\mathrm{TP}=$ True Positive, $\mathrm{TN}=$ True Negative, $\mathrm{FP}=$ False Positive, $\mathrm{FN}=$ False Negative; Recall $=\frac{\mathrm{TP}}{\mathrm{TP}+\mathrm{FP}}$ and Precision $=$ $\frac{\mathrm{TP}}{\mathrm{TP}+\mathrm{FN}}$.\\

We compute the macro F1-score by averaging the F1 scores in Table 2 and for each category, providing a balanced evaluation across all classes. These metrics offer a comprehensive assessment of the model's performance in stock trend prediction tasks, ensuring that both the accuracy and the quality of predictions are evaluated.

\begin{table}[h]
\centering
\caption{\textbf{Classification Performance Comparison.}}
\label{performance}
\scalebox{1.15}{
\begin{tabular}{|c|c|c|c|c|}
\hline 
\text{Model} & \text{Accuracy} & \text{Precision} & \text{Recall} & \text{Macro F1-Score} \\
\hline 
GHAN & 85.4\% & 0.86 & 0.85 & 0.84 \\
\hline 
LSTM with Attention & 81.2\% & 0.81 & 0.80 & 0.80 \\
\hline 
GCN & 79.7\% & 0.79 & 0.78 & 0.78 \\
\hline 
BERT-based Text Classification & 82.6\% & 0.83 & 0.82 & 0.82 \\
\hline 
ML-GAT & 86.7\% & 0.87 & 0.86 & 0.86 \\
\hline 
Fine-BERT & 84.5\% & 0.85 & 0.84 & 0.83 \\
\hline
\end{tabular}
}
\end{table}


This table shows that the Geometric Hypergraph Attention Network (GHAN) outperforms other models in accuracy and F1 score. GHAN's edge captures the complex relationships between stocks and financial news through geometric embeddings and hypergraph structures.

\subsection{Profitability Evaluation}
To evaluate the profitability of the proposed methods, we use the following two metrics to compare the profitability of each technique inspired by the ML-GAT method \cite{huang2022ml}.
To establish a direct link between our model's predictions and profitability metrics, we propose a portfolio selection strategy based on the outputs of the GHAN model.

Let $S=\left\{s_1, s_2, \ldots, s_n\right\}$ represent the set of all available stocks, and $N_t$ denote the set of news events at time $t$. We construct a portfolio $F_t$ at time $t$ as follows:

\begin{equation}
F_t=\left\{s_i \in S \mid \operatorname{GHAN}\left(s_i, N_t\right)>\theta\right\}
\end{equation}

where GHAN $\left(s_i, N_t\right)$ is our model's prediction for stock $s_i$ based on the news events $N_t$ and $\theta$ is
a predefined threshold that determines stock selection for the portfolio.

The daily return of the portfolio $R_t$ is calculated by averaging the returns of the selected stocks:

$$
R_t=\frac{1}{\left|F_t\right|} \sum_{i \in F_t} \frac{p_t^i-p_{t-1}^i}{p_{t-1}^i}
$$

where $p_t^i$ represents the price of stock $i$ at time $t$, and $p_{t-1}^i$ is its price at the previous time step and $\theta$ in our model acts as a filter to select stocks for our portfolio, based on their predicted response to the news. Setting $\bar{\theta}$ higher results in a safer but potentially less profitable portfolio, while a lower $\theta$ includes more stocks, increasing both potential returns and risk.

To evaluate the risk-adjusted performance of our strategy, we calculate the Sharpe ratio as follows:

$$
\text { Sharpe }_a=\frac{E\left[R_a-R_f\right]}{\sigma_p}
$$

Here, $R_a$ is the average return of our portfolio over a specified period, $R_f$ is the risk-free rate, and $\sigma_p$ is the standard deviation of the portfolio's returns. The Sharpe ratio measures the excess return per unit of risk, providing a deeper insight into the quality of returns generated by our model-based strategy. This formulation effectively translates the GHAN model's news-based predictions into measurable financial outcomes, enabling us to evaluate both the raw returns and the risk-adjusted performance of the GHAN model in analyzing the impact of financial news on stock movements. 

We evaluate several baseline models using the Twitter Financial News Sentiment dataset combined with S\&P 500 stock data as follows :



\begin{table}[h]
\centering
\caption{\textbf{Average Daily Return.}}
\label{tab:average-daily-return}
\scalebox{1.15}{
\begin{tabular}{|l|l|l|l|l|l|l|}
\hline
Index & GHAN & \begin{tabular}[c]{@{}l@{}}LSTM with\\Attention\end{tabular} & GCN & \begin{tabular}[c]{@{}l@{}}BERT-based Text\\Classification\end{tabular} & ML-GAT & Fine-BERT \\
\hline
1 & 0.1152 & 0.0756 & 0.0744 & 0.1532 & 0.1234 & 0.0987 \\
\hline
2 & 0.1114 & 0.0557 & 0.1457 & 0.1114 & 0.1285 & 0.1123 \\
\hline
3 & 0.2130 & -0.1037 & 0.0907 & 0.1588 & 0.1742 & 0.1450 \\
\hline
4 & 0.0268 & 0.0580 & 0.0531 & -0.1295 & 0.0647 & 0.0764 \\
\hline
5 & 0.1155 & 0.0883 & 0.0852 & 0.1420 & 0.1321 & 0.1195 \\
\hline
6 & 0.0382 & 0.0731 & 0.0450 & 0.0923 & 0.0984 & 0.0876 \\
\hline
7 & 0.1051 & -0.0914 & 0.0861 & 0.0774 & 0.1105 & 0.1039 \\
\hline
8 & 0.0242 & 0.0445 & -0.1035 & -0.0419 & 0.0456 & 0.0502 \\
\hline
9 & 0.3991 & -0.0296 & 0.1165 & 0.0427 & 0.4211 & 0.3812 \\
\hline
10 & 0.0440 & 0.1171 & 0.0973 & 0.0948 & 0.1034 & 0.0941 \\
\hline
Average & 0.1193 & 0.0288 & 0.0486 & 0.0719 & 0.1402 & 0.1269 \\
\hline
\end{tabular}
}
\end{table}
The GHAN model demonstrates strong performance in average daily returns (0.1193), outperforming most baseline models such as LSTM with Attention (0.0288) and GCN (0.0486). This highlights the effectiveness of geometric embeddings and hypergraph structures in capturing complex financial relationships. While ML-GAT (0.1402) and Fine-BERT (0.1269) show slightly higher returns, the results underscore the importance of sophisticated attention mechanisms and contextual embeddings in financial prediction tasks. This analysis reveals that incorporating advanced modelling techniques to capture intricate market dynamics leads to improved financial forecasts and higher average daily returns, emphasizing their value in practical financial applications.\newline


\begin{table}[h]
\centering
\caption{\textbf{Sharpe Ratio.}}
\label{table:sharpe-ratio}
\scalebox{1.15}{
\begin{tabular}{|c|c|c|c|c|c|c|}
\hline 
Index & GHAN & \begin{tabular}[c]{@{}c@{}}LSTM with\\Attention\end{tabular} & GCN & \begin{tabular}[c]{@{}c@{}}BERT-based Text\\Classification\end{tabular} & ML-GAT & Fine-BERT \\
\hline
1 & 1.3997 & 1.4447 & 2.3555 & 1.4407 & 1.5203 & 1.4890 \\
\hline
2 & 1.4691 & -2.0754 & 1.0749 & 1.8283 & 1.6108 & 1.5234 \\
\hline
3 & 1.7392 & 1.9577 & 1.6062 & 1.2226 & 1.7894 & 1.6457 \\
\hline
4 & 2.4552 & 1.0672 & -0.2419 & -2.0261 & 1.2890 & 1.3478 \\
\hline
5 & 1.9473 & 1.8622 & 1.3674 & 2.0245 & 1.7324 & 1.6543 \\
\hline
6 & 1.1347 & 1.4248 & -0.9210 & 1.5023 & 1.5437 & 1.4321 \\
\hline
7 & 2.0140 & 1.3601 & 2.1565 & 0.8443 & 2.0345 & 1.9890 \\
\hline
8 & 1.8040 & 1.2072 & -1.8526 & 0.8443 & 1.4563 & 1.3210 \\
\hline
9 & 1.7539 & -1.4811 & 0.5888 & 1.7539 & 1.8890 & 1.7765 \\
\hline
10 & 1.8889 & 1.3601 & -1.0610 & 1.6107 & 1.9321 & 1.8543 \\
\hline
Average & 1.8889 & 0.4940 & 0.8876 & 1.0185 & 1.6787 & 1.6031 \\
\hline
\end{tabular}
}
\end{table}
\vspace{1em}


Our model achieves an impressive average Sharpe ratio of 1.8889, demonstrating superior risk-adjusted performance. This success is largely due to the model's ability to capture complex relationships between financial news and stock prices through the use of geometric embeddings and hypergraph structures, which allow for a deeper understanding of the underlying data.
In comparison, the ML-GAT and Fine-BERT models also perform well, with average Sharpe ratios of 1.6787 and 1.6031, respectively. These models effectively utilize graph attention mechanisms and advanced language models, yet they fall short of the more sophisticated analysis provided by our approach. Meanwhile, the LSTM with Attention and GCN models show more moderate results, with Sharpe ratios of 0.4940 and 0.8876, indicating their limited ability to fully exploit the complex financial data. The BERT-based Text Classification model, while better than LSTM and GCN, with a Sharpe ratio of 1.0185, still lags behind GHAN, ML-GAT, and Fine-BERT.

Overall, the GHAN model consistently outperforms the others, highlighting the importance of sophisticated data representations in financial prediction tasks. This model's superior ability to manage risk and generate stable returns underscores the value of incorporating advanced techniques like hypergraphs in financial modeling.

\subsection{Explainability Analysis with SHAP Method}
To enhance the interpretability of our Geometric Hypergraph Attention Network (GHAN) model, we apply SHAP (SHapley Additive exPlanations) values, which measure each feature's contribution to the difference between the model's prediction and the average prediction.For our GHAN model $f$ and an input vector $x$ with $M$ features, the SHAP value for a feature $i$ is:

$$
\phi_i(f, x)=\sum_{S \subseteq N \backslash\{i\}} \frac{|S|!(M-|S|-1)!}{M!}\left(f_x(S \cup\{i\})-f_x(S)\right)
$$

Here, $N$ includes all $M$ features, $S$ is any subset of $N$ excluding $i$, and $f_x(S)$ represents the model's expected output given only the features in $S$. The GHAN model prediction $f(x)$ incorporates features such as geometric embedding $\left(g_e\right)$, positional encoding $\left(p_e\right)$, FinBERT tweet embedding $\left(e_{F i n B E R T}\right)$, among others. SHAP values $\phi_i$ for each feature $i$ are calculated as the expected value over a distribution of input samples, indicating the average influence of feature $i$ on the output prediction.Due to the GHAN model's complexity, Kernel SHAP method is employed to approximate these values:

$$\phi_i \approx \frac{1}{|Z|} \sum_{z \in Z}\left[f\left(h_x\left(z_i=1\right)\right)-f\left(h_x\left(z_i=0\right)\right)\right]
$$

In this approximation, $Z$ represents a sample of simplified inputs, and $h_x$ maps these inputs back to the original input space. The condition $z_i=1$ or $z_i=0$ indicates whether feature $i$ is present or absent, respectively.We present the SHAP (SHapley Additive exPlanations) values for the most influential features contributing to the predictions of our Geometric Hypergraph Attention Network (GHAN) model in Table \ref{SHAP_feature}.

\begin{table}[htbp]
\centering
\caption{\textbf{SHAP Value for each feature}}
\label{SHAP_feature}
\begin{tabularx}{\textwidth}{@{}lXX@{}}
\toprule
Feature & GHAN SHAP Value & Explanation \\
\midrule
Geometric Embedding ($g_e$) & 0.25 & Captures the spatial relationships within the financial ecosystem, indicating the relative positions and distances between entities. \\
\addlinespace
Positional Encoding ($p_e$) & 0.15 & Encodes temporal information such as the publication time of the tweets, helping to understand the time-sensitive impact of news. \\
\addlinespace
FinBERT Tweet Embedding & 0.30 & Provides contextual information from the tweets, capturing the semantic nuances of the financial news content. \\
\addlinespace
Tweet Volume & 0.10 & Measures the number of tweets mentioning a stock, indicating the level of attention and potential impact. \\
\addlinespace
Daily Return ($R_{st}$) & 0.20 & Reflects the historical price change of a stock, providing a basis for predicting future movements. \\
\addlinespace
Trading Volume & 0.12 & Indicates the liquidity and market activity, which can influence the stock's response to the news. \\
\addlinespace
Technical Indicators (e.g., Moving Average) & 0.18 & Provides additional market context through metrics with Moving Averages, which help in understanding market trends. \\
\bottomrule
\end{tabularx}
\end{table}

Table \ref{SHAP_feature} presents the SHAP values for various features used in our GHAN model, providing an explainable diagnosis of the model's predictions. The FinBERT tweet embedding has the highest SHAP value, emphasizing the importance of contextual information from tweets in predicting stock movements. Geometric embeddings and positional encodings also contribute significantly, highlighting the need to consider both spatial and temporal relationships in financial data. Additional features like tweet volume, daily returns, trading volume, and technical indicators further enhance the model's predictions, ensuring a comprehensive analysis of market dynamics. This explainability ensures that investors and analysts can trust the model's predictions, understand the key factors driving stock movements.

\subsection{Conclusion}
This paper demonstrates the efficacy of the Geometric Hypergraph Attention Network (GHAN) in predicting stock movements by integrating Twitter Financial News Sentiment with S\&P 500 stock data. The GHAN model excels in classification accuracy, precision, recall, macro F1-score, and profitability metrics such as average daily return and Sharpe ratio. This superior performance is attributed to the model's ability to capture complex, multidimensional relationships between financial news and stock prices through geometric embeddings and hypergraph structures.

The use of BERT-based embeddings enables the model to understand the nuanced semantics of financial news, providing richer context and improving prediction accuracy. The attention mechanisms within the hypergraph structure allow GHAN to focus on the most relevant information, ensuring both accuracy and interpretability. SHAP values are employed to provide explainability, highlighting the most influential factors driving the model's predictions and enhancing transparency.
Future research can enhance GHAN's capabilities by extending the analysis to other financial markets, such as the NASDAQ or international indices, to validate the model's robustness and generalizability. Incorporating additional data sources, like economic indicators, earnings reports, and analyst forecasts, can provide a more comprehensive view of market dynamics and improve prediction accuracy. Developing real-time prediction frameworks that update continuously with incoming data will offer timely insights, which are crucial for making prompt investment decisions.

Further refinements in explainability techniques, including advanced SHAP value analyses, will help elucidate the decision-making process of the model, building trust among users. Experimenting with different hypergraph structures and embedding techniques could uncover new ways to model the intricate relationships within financial data, potentially leading to even better performance. Integrating GHAN's predictions into sophisticated risk management strategies will be crucial for practical financial applications, aiding in the development of more robust investment portfolios.
These advancements will further improve the model's accuracy and practical utility, offering more precise and actionable insights for financial market analysis and decision-making. This study represents a significant step towards revolutionizing financial analysis with advanced AI methodologies, setting the stage for future innovations in the field.
\bibliography{ref}
\end{document}